\pdfoutput=1

\documentclass[11pt]{article}

\PassOptionsToPackage{dvipsnames}{xcolor}
\usepackage{hyperref}
\usepackage[x11names, svgnames, table]{xcolor}

\definecolor{aclpurple}{HTML}{800080} % hex for purple

\hypersetup{
    colorlinks=true,
    citecolor=aclpurple,   % citations
    linkcolor=aclpurple,   % internal links
    urlcolor=aclpurple     % URLs, including mailto links
}

\usepackage[final]{assets/acl}   % For final/camera-ready
\usepackage{times}
\usepackage{latexsym}
\usepackage[T1]{fontenc}   % Proper rendering of Latin characters
\usepackage[utf8]{inputenc} % UTF-8 encoding
\usepackage{inconsolata}   % Better typewriter font
\usepackage{graphicx}      % Include images
\usepackage{expex}

\usepackage{tipa}
\usepackage{tikz}
\usepackage{makecell}
\usepackage{rotating}
\usepackage{amssymb}
\usepackage{pifont} 
\usepackage{tabularray} 
\usepackage{array}
\usepackage{fdsymbol}
\usepackage{cleveref}
\usepackage{multirow}
\usepackage[scaled=0.85]{helvet}
\usepackage{tcolorbox}
\usepackage{xcolor}
\usepackage{xspace}
\usepackage{booktabs}
\usepackage{tabularx}
\usepackage[table]{xcolor}
\usepackage{caption}
\usepackage{subcaption}
\usepackage{setspace}
\usepackage{cellspace}
\usepackage{twemojis}

\usepackage{linguex}
\usepackage{pgfplots}
\pgfplotsset{compat=1.18} % or the version installed

\renewcommand{\arraystretch}{1.05}
\newcolumntype{Y}{>{\raggedright\arraybackslash}X}

\newcommand{\cellcomp}[2]{%
  \pgfmathparse{#1-#2}%
  \ifdim\pgfmathresult pt>0pt
    \cellcolor{green!18}{#1}%
  \else
    \cellcolor{red!15}{#1}%
  \fi
}

\makeatletter
\DeclareRobustCommand*{\escapeus}[1]{%
  \begingroup\@activeus\scantokens{#1\endinput}\endgroup}
\begingroup\lccode`\~=`\_\relax
\lowercase{\endgroup\def\@activeus{\catcode`\_=\active \let~\_}}
\makeatother
\newcommand{\myemph}[1]{\textsf{{\escapeus{#1}}}}

\newcommand\cincludegraphics[2][]{\raisebox{-0.3\height}{\includegraphics[#1]{#2}}}

\usepackage{listings}
\usepackage{graphicx}   % optional, often for images
\usepackage{caption}    % optional, improves captions

\newif\ifshowemoji
\showemojitrue   % comment this line out in camera-ready / arXiv

\newcommand{\bemoj}[1]{\ifshowemoji\texttwemoji{#1}\fi}

\title{%
    \centering
    BLiSS 1.0: Evaluating Bilingual Learner Competence in Second Language Small Language Models
}

\author{
    {\bf Yuan Gao}\thanks{\textbf{Corresponding Authors:  \href{mailto:yg386@cam.ac.uk}{yg386@cam.ac.uk}, 
\href{mailto:sas245@cam.ac.uk}{sas245@cam.ac.uk}}
}~\bemoj{hatching_chick}\bemoj{baby_chick} \quad 
    {\bf Suchir Salhan*}~\bemoj{hatching_chick}\bemoj{baby_chick} \quad
    {\bf Andrew Caines}~\bemoj{hatching_chick}\bemoj{baby_chick} \\ 
    {\bf Paula Buttery}~\bemoj{hatching_chick}\bemoj{baby_chick} \quad
    {\bf Weiwei Sun}~\bemoj{baby_chick} \\
\bemoj{hatching_chick} ALTA Institute \bemoj{baby_chick} Department of Computer Science \& Technology, University of Cambridge
}

\begin{document}

\maketitle

% ========
% ABSTRACT
% ========
\begin{abstract}
To bridge the gap between performance-oriented benchmarks and the evaluation of cognitively-inspired models, we introduce \textbf{BLiSS 1.0}, a Benchmark of Learner Interlingual Syntactic Structure. Our benchmark operationalizes a new paradigm of selective tolerance, testing if a model finds a naturalistic learner error more plausible than a matched, artificial error within the same sentence. Constructed from over 2.8 million naturalistic learner sentences, BLiSS provides 136,867 controlled triplets (corrected, learner, artificial) for this purpose. Experiments on a diverse suite of models demonstrate that selective tolerance is a distinct capability from standard grammaticality, with performance clustering strongly by training paradigm. This validates BLiSS as a robust tool for measuring how different training objectives impact a model's alignment with the systematic patterns of human language acquisition.
\end{abstract}

\noindent

\noindent
\begin{tblr}{colspec = {Q[c,m]|X[l,m]}, stretch = 0}
    % First row: Hugging Face logo + link/text
    \cincludegraphics[width=1.35em, keepaspectratio]{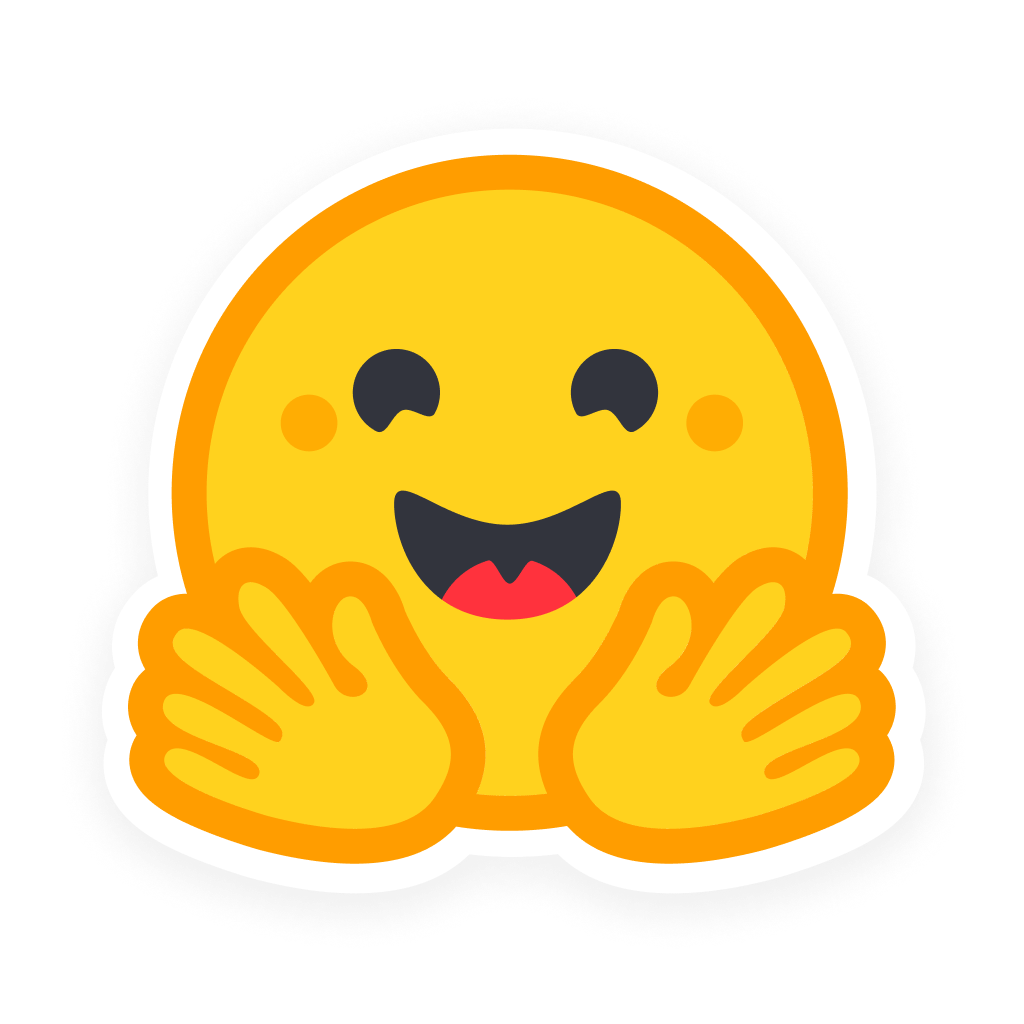} &
    \setstretch{.5}{\small\myemph{\textbf{BLiSS} on \href{https://huggingface.co/ALTACambridge}{HuggingFace} (BLiSS 1.0 Dataset and Pretrained Models)}} \\

    % Second row: GitHub logo + link/text
    \cincludegraphics[width=1.1em, keepaspectratio]{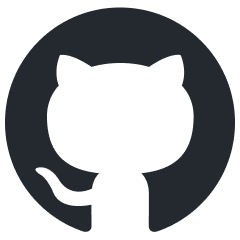} &
    \setstretch{.5}{\small\myemph{Training Code Open-Sourced on \href{https://github.com/yuan-w-gao/bliss}{GitHub}}}
\end{tblr}

% ============
% INTRODUCTION
% ============
\section{Introduction}
\label{sec:intro}
There is a growing interest in the NLP community in developing models that are not just powerful, but also cognitively inspired—that is, models which aim to reflect the processes of human language acquisition.
Current evaluation benchmarks for language models are overwhelmingly performance-oriented, centering around grammaticality tests, adherence to standard grammar, and task performance  (e.g., BLiMP \citet{warstadt-etal-2020-blimp-benchmark} and GLUE \citet{wang-etal-2018-glue}).
While these measures are informative in evaluating linguistic competence, the core question for cognitively inspired modeling is different: do our systems exhibit the kinds of behaviors that emerge in human acquisition?
%If not, higher accuracy only makes a model better at being unlike us. 
For models that aim to be cognitively plausible, we need a complementary, acquisition-focused perspective, one that inspects how grammar competence is organized and learned. 

This evaluation gap is particularly important for models of Second Language Acquisition (SLA), which we refer to as L2LMs \citep{aoyama2024ModelingNonnativeSentence}. 
A central characteristic of the SLA process is the production of systematic `errors'. 
These deviations are not random noise, but rather structured evidence of the learner's developing internal grammar, or "interlanguage" \citep{corder2015significance, Selinker+1972+209+232}.
For a model to be truly `learner-like', it must be sensitive to these specific, structured patterns observed in real human data. 

% Operationalizing this sensitivity, however, presents a significant methodological challenge. A straightforward minimal-pair test, comparing a model’s preference for a learner sentence over its corrected version, is conceptually problematic. It directly confounds learner-like behavior with a general failure to model the target grammar, as a model could achieve a high score simply by being ungrammatical. Therefore, a more robust paradigm is required—one that can disentangle sensitivity to learner patterns from overall grammatical competence.

To address this, we propose a new paradigm built on a key assumption: the systematicity of learner errors is tied to both the type of error and its specific locus within a sentence. 
This assumption, therefore, theorizes that moving an attested error to a different, albeit plausible, location renders it less naturalistic and less human-like. 
This approach, which uses an error's locus as a test of naturalness, is inspired by similar methodologies for evaluating complex linguistic phenomena \citep{sterner-teufel-2025-minimal}. 
This allows us to test a model's selective tolerance: its ability to penalize a naturalistic human error less severely than a contrived artificial-locus error of the same type. 

We introduce the \textbf{Benchmark of Learner Interlingual Syntactic Structure (BLiSS 1.0)}, a large-scale evaluation dataset on a model's alignment with naturalistic language learner patterns, offering a new dimension of evaluating acquisition-focused models. 
BLiSS is built upon three of the largest English learner corpora available: the EF-Cambridge Open Language Database (EFCAMDAT) \citep{Geertzen2014AutomaticLA}, the Write \& Improve Corpus (W\&I) \cite{wicorpus24}, and the First Certificate in English (FCE) dataset \citep{yannakoudakis-etal-2011-new}.

% \begin{exe}
% \ex U:DET (Unnecessary determiner)
% \label{citko21}
% \begin{xlist}
%  \ex {There are a lot of benefits when we play sports.}
%  \ex[*] {There are a lot of benefits when we play \textbf{the} sports.}
%  \ex[**]{There are a lot of benefits when \textbf{the we} play sports.}
% \end{xlist}
% \end{exe}
\ex. U:DET (Unnecessary determiner) \label{citko21}
\a. There are a lot of benefits when we play sports.
\b. *There are a lot of benefits when we play \textbf{the} sports.
\c. **There are a lot of benefits when \textbf{the} we play sports.

The core of the BLiSS evaluation is the triplet, a controlled comparison between: a corrected sentence, the original sentence with one error from a learner, and a version with an artificially-generated error of the same error type, as shown in \ref{citko21}. 
From an initial pool of over 2.8 million raw learner sentence-corrected sentence pairs, we systematically generate a matched artificial-locus for each valid, single-edit grammatical deviation. 
After a rigorous multi-stage validation pipeline, BLiSS comprises 136,867 high-quality evaluation triplets. 
Each triplet is accompanied by rich metadata, including learner L1, proficiency level, and error type, as illustrated in Figure~\ref{fig:udet}. 

In this paper, we deploy BLiSS to evaluate a diverse suite of models, from large bilingual LLMs to acquisition-inspired L2LMs. Our results yield two key findings. First, we demonstrate that selective tolerance is a distinct capability from standard grammaticality; high performance on BLiMP does not guarantee high performance on BLiSS. Second, we show that model performance on BLiSS clusters strongly by training paradigm, validating it as a tool for measuring how different architectures and training objectives impact a model's alignment with the systematic patterns of human learner language.

\begin{figure}[ht]
\centering
\begin{minipage}{0.95\linewidth}
\lstdefinelanguage{json}{
    basicstyle=\ttfamily\small,
    numbers=left,
    numberstyle=\tiny\color{gray},
    stepnumber=1,
    numbersep=5pt,
    showstringspaces=false,
    breaklines=true,
    backgroundcolor=\color{gray!5},
    literate=
     *{0}{{{\color{blue}0}}}{1}
      {1}{{{\color{blue}1}}}{1}
      {2}{{{\color{blue}2}}}{1}
      {3}{{{\color{blue}3}}}{1}
      {4}{{{\color{blue}4}}}{1}
      {5}{{{\color{blue}5}}}{1}
      {6}{{{\color{blue}6}}}{1}
      {7}{{{\color{blue}7}}}{1}
      {8}{{{\color{blue}8}}}{1}
      {9}{{{\color{blue}9}}}{1}
      %{:}{{{\color{red}{:}}}}{1}
      %{,}{{{\color{red}{,}}}}{1}
      {"}{{{\color{black}{"}}}}{1}
      {the}{{\textbf{\color{blue}the}}}3
      {we}{{\textbf{\color{red}we}}}2
      {play}{{\textbf{\color{OliveGreen}play}}}4
      {sports}{{\textbf{\color{OliveGreen}sports}}}6
}

\begin{lstlisting}[language=json]
{
  "learnerID": "8421",
  "L1": "Vietnamese",
  "cefr": "C1",
  "topic": "play sports",
  "corrected": "There are a lot of benefits when we play sports.",
  "learner error": "There are a lot of benefits when we play the sports.",
  "artificial error": "There are a lot of benefits when the we play sports.",

  "errant_edits": [{
    "type": "U:DET",
    "o_str": "the",
    "c_str": ""
  }],
  "all_error_types": [
    "U:DET"
  ]
}
\end{lstlisting}

\end{minipage}
\caption{An example BLiSS triplet illustrating an Unnecessary Determiner (U:DET) error. The original learner sentence contains an unnecessary determiner “the”, which is removed in the corrected sentence. Artificially-generated errors of the same type allow controlled evaluation of model preferences.}
\label{fig:udet}
\end{figure}

\section{Related Work}
\subsection{Second Language Acquisition-Inspired Language Models (L2LMs)}
We use L2LMs to denote cognitively inspired models of L2 acquisition \citep{aoyama2024ModelingNonnativeSentence}. Early work examined transfer—training on an L1 then an L2—and the role of typological distance \citep{yadavalli-etal-2023-slabert, oba-etal-2023-second}, while later studies add cognitive priors (e.g., alignment to learner reading times; preserving L1 knowledge to probe the Critical Period Hypothesis) and compare sequential vs.\ mixed L1/L2 exposure \citep{aoyama2024ModelingNonnativeSentence, clahsen2006a, constantinescu2025InvestigatingCriticalPeriod, Lenneberg1967biological, kirkpatrick2017overcoming, Arnett2025OnTA}. Given heterogeneous architectures and pretraining corpora (from learner-like data to web-scale sources such as CC-100; \citealp{wenzek2020ccnet}), a common benchmark tied to learner behavior is needed \citep{salhan-etal-2024-less, Arnett2025OnTA}.
\subsection{Learner Corpora and Error Profiling }

Large-scale learner corpora provide an important empirical basis for modeling and evaluating L2 learner behavior. The Write \& Improve Corpus 2024 \citep{wicorpus24} contains learner essays with Common European Framework of Reference for Languages (CEFR) annotations and corresponding error-labeled corrections. The essays were submitted by users of the `Write \& Improve' writing practice platform\footnote{\url{https://writeandimprove.com/}}. W\&I uses ERRANT \citep{bryant-etal-2017-automatic} to annotate errors in learner essays automatically. ERRANT annotations classify errors as replacements (\texttt{R}), missing (\texttt{M}) or unnecessary (\texttt{U}) and assign a specific tag (e.g., \texttt{M:ADJ} means the text omits an adjective). A full table of ERRANT error codes are included in \textit{Appendix} \ref{errant} for reference.  The EF-Cambridge Open Language Database (EFCAMDAT) \citep{Geertzen2014AutomaticLA} offers a large collection of learner texts annotated with proficiency levels and metadata on learner nationality. Note that the proficiency levels in EFCAMDAT relate to difficulty level attained by users of the `EF Englishtown' platform (now `EF English Live'\footnote{\url{https://englishlive.ef.com/}}), rather than human ratings of the texts themselves, but this information serves as a good proxy for learner proficiency. The W\&I-2024 corpus has a wider range of L1s compared to other publically-available learner corpora, like the FCE subset of the Cambridge Learner Corpus \citep{yannakoudakis-etal-2011-new}. There are other error-annotated English learner corpora, such as NUCLE \citep{dahlmeier2013nus}, JFLEG \citep{napoles2017jfleg} and Lang-8 \citep{ mizumoto2012effect, tajiri2012tense}, but are respectively age/language restricted; use fluency rewrite rather than minimal grammatical edits; and have user-generated corrections \citep{wicorpus24}. 

\section{BLiSS 1.0}
\label{sec:bliss_data}

\subsection{Motivation}

The BLiSS 1.0 benchmark is a large-scale evaluation suite composed of controlled triplets designed to test a model's selective tolerance for naturalistic learner production errors. The evaluation framework for BLiSS is designed to move beyond evaluations of the formal competence of a Language Model (e.g., using broad-coverage datasets like BLiMP \citep{warstadt-etal-2020-blimp-benchmark}) to evaluate the \textbf{alignment of a language model with second language acquisition}. BLiSS builds upon previous attempts to extend acquisition-inspired evaluation frameworks for Language Models (e.g., \citet{evanson2023language}) beyond first language acquisition. 

BLiSS 1.0 focuses on naturalistic production errors in learner corpora. The BLiSS 1.0 benchmark is designed to evaluate how closely a language model’s outputs align with patterns observed in second language (L2) learners, particularly in terms of grammatical errors. While it is true that individual learner errors do not imply that a majority of learners would make the same mistake in a given sentence, BLiSS focuses on systematic tendencies in learner language rather than absolute probabilities of specific errors. By aggregating errors across millions of sentence-correction pairs from multiple learner corpora, BLiSS captures the distributional patterns of learner errors that are prevalent in naturalistic L2 production. BLiSS does not encourage models to prefer errors, but rather tests alignment with learner error patterns.

This approach addresses a critical limitation of traditional LM evaluation benchmarks (e.g., BLiMP), which primarily assess formal grammatical competence. Such benchmarks assume that the model should always prefer grammatical sentences, but human learners – especially in L2 acquisition –frequently produce systematic errors that reveal underlying acquisition stages, interlanguage phenomena, or L1 transfer effects. BLiSS thus extends evaluation beyond formal competence, providing a framework to test whether models \textit{selectively tolerate or reproduce error patterns} in ways that resemble human learners.

Concretely, we develop BLiSS to enable the study of:

\begin{enumerate}
    \item \textbf{Error-type sensitivity:} Whether language models recognize and react differently to common L2 errors (e.g., determiner omission, verb tense errors).
    \item \textbf{Position awareness:} By generating artificial errors at positions distinct from the learner’s original error, we can test if language models are sensitive to the locus of grammatical deviations, not just their existence.
    \item \textbf{Learner-informed evaluation:} Leveraging metadata such as L1 background and proficiency level that are available in large-scale corpora allows analysis of model behavior in the context of typologically diverse learner populations.
\end{enumerate}

While BLiSS does not imply that \textit{all learners} would produce a given error, it provides a systematically sampled and validated set of errors that represents frequent phenomena in learner production \citep{alexopoulou2015exploring, le2021learner, crossley202234, alexopoulou2022big}. This makes BLiSS a meaningful benchmark for probing the alignment of language models with human L2 acquisition patterns, without conflating individual idiosyncrasies with population-level tendencies.

% The primary focus of this work is on BLiSS-Precise, a dataset constructed for highly controlled analysis, with rigorous quality checks in place.  
% We also provide BLiSS-Coverage, a dataset maximizing the diversity of linguistic phenomena and coverage. The size difference between the two subsets is relatively small with 121,479 triplets in BLiSS-Precise and 148,410 in BLiSS-Coverage. 
% Therefore, the remainder of this section and the following analysis will focus primarily on the more carefully curated BLiSS-Precise dataset. 

\subsection{Data: Source Corpora}
The credibility and naturalistic grounding of the BLiSS benchmark stem from its foundation in large-scale, naturalistic learner data. 
We aggregate sentence-correction pairs from three of the most widely-used English learner corpora, ensuring our benchmark reflects genuine learner behaviour in communicative contexts. 
\begin{itemize}
    \item The EF-Cambridge Open Language Database (EFCAMDAT) \citep{Geertzen2014AutomaticLA}: A very large collection of over 1 million learner texts from an online English learning platform. Texts are annotated with metadata including learner nationality and proficiency levels mapped to the Common European Framework of Reference for Languages (CEFR).
    \item The Write \& Improve (W\&I) Corpus \citep{wicorpus24}: A dataset of learner essays submitted to an online writing feedback tool. It is richly annotated with CEFR levels (A1–C2) and explicit learner L1 labels, providing high-quality metadata for fine-grained analysis.
    \item The First Certificate in English (FCE) Dataset \citep{yannakoudakis-etal-2011-new}: A well-known subset of the Cambridge Learner Corpus containing essays from an official language proficiency exam. This provides a valuable sample of argumentative, exam-style writing from a diverse set of L1 backgrounds.
\end{itemize}
Collectively, these corpora provide a massive pool of over 2.8 million raw sentence-correction pairs, forming the empirical starting point for our triplet construction pipeline, detailed in the following section.

\begin{table}[htbp]
  \centering
  \begin{tabular}{@{}l|l@{}}
    Corpus & \# Raw Pairs \\
    \midrule
    \midrule
    EFCAMDAT & 2,711,188 \\
    W\&I & 63,926 \\
    FCE & 52,421 \\
    \midrule 
    Total & 2,827,535 \\
  \end{tabular}
  \caption{Summary of raw single-edit sentence-correction pairs from the source corpora.}
  \label{tab:rawpair}
\end{table}

\subsection{Triplet Construction Pipeline}
The construction of the BLiSS dataset follows a multi-stage pipeline designed to transform raw sentence-correction pairs from the source corpora into high-quality, validated triplets. 
The pipeline emphasizes grammatical precision, methodological transparency, and the atomization of errors to ensure each triplet tests a single distinct linguistic phenomenon. 

\paragraph{Grammatical Error Classification and Filtering}
The process begins with a comprehensive error analysis of the raw sentence pairs using the ERRANT toolkit \citep{bryant-etal-2017-automatic}. 
With these annotations, we first filtered out pairs containing only non-grammatical edits, such as spelling, punctuation, or capitalization changes. 

\paragraph{Error Atomization}
We then atomized sentence pairs with multiple corrections using the ERRANT annotations as a guide. 
Each distinct grammatical edit within a multi-error sentence was isolated to create a new single-edit pair consisting of the corrected sentence and a version with just that one specific error. 
This process ensures that every triplet in the final dataset is anchored to exactly one grammatical deviation, allowing for a clean and targeted evaluation. 

\paragraph{Rule-Based Artificial Error Generation}
The core of the pipeline is the generation of an artificial error for each single-edit pair. 
This rule-based system uses linguistic analysis and morphological generation, creating a new sentence that adheres to two fundamental constraints:
\begin{enumerate}
    \item \textbf{Error Type Consistency: }the artificial error must mirror the grammatical operation of the human error. For example, a missing determiner (M:DET) in the learner sentence prompts the generation of a new sentence where a determiner is removed. 
    \item \textbf{Position Divergence:} The artificial error must be introduced at a different word position than the learner error. This ensures the model is being tested on its sensitivity to the error's locus, not merely its presence. 
\end{enumerate}

\paragraph{Multi-Stage Quality Validation}
To ensure the integrity of BLiSS, every generated triplet was subjected to a rigorous multi-stage validation filter. A triplet was only retained if it passed all of the following checks:
\begin{enumerate}
    % \item Syntactic Validity: The generated sentence must be parsable by spaCy \footnote{https://spacy.io/}. 
    \item \textbf{Morphological Correctness: }All inflected words (e.g., verbs, nouns) generated by LemmInflect\footnote{https://github.com/bjascob/LemmInflect} must be valid English forms. 
    \item \textbf{Triplet Uniqueness: }The artificial error sentence must be distinct from both the corrected sentence and the original learner error sentence. 
    \item \textbf{Error Type Confirmation:} Finally, we used ERRANT as a verifier. The generated artificial error, when compared to the corrected sentence, must be classified by ERRANT as having the same error type as the original human error. 
\end{enumerate}
This stringent validation process resulted in an overall success rate of 4.8\%, yielding a final dataset of 136,867 high-quality triplets. The low success rate is a direct reflection of the strictness of our quality controls, ensuring that every item in BLiSS is a valid and non-ambiguous test case. A sample of 100 triplets was also manually reviewed, confirming a grammatical and positional accuracy rate of over 95\%.

\subsection{Dataset Composition}
Following the rigorous construction and validation pipeline, the final BLiSS dataset comprises 136,867 high-quality triplets. 
The composition of the dataset reflects both the diversity of the source corpora and the targeted nature of our filtering process. 
As shown in Table~\ref{tab:corpus-triplets}, the majority of the final dataset (76.7\%) is derived from the large-scale EFCAMDAT corpus, supplemented by high-quality and diverse data from the W\&I and FCE corpora. 

\begin{table}[htbp]
  \centering
  \begin{tabular}{@{}lrr@{}}
    \toprule
    Corpus & Triplets & Percentage \\
    \midrule
    EFCamDat & 105,034 & 76.7\% \\
    Write \& Improve & 17,380 & 12.7\% \\
    FCE & 14,453 & 10.6\% \\
    \midrule
    Total & 136,867 & 100\% \\
    \bottomrule
  \end{tabular}
  \caption{BLiSS Composition}
  \label{tab:corpus-triplets}
\end{table}

\paragraph{Error Type Distribution}
The dataset provides robust coverage across a range of core grammatical error categories that are common in second language acquisition. 
Table~\ref{tab:errant-top5-bliss-precise} details the distribution of the five most frequent error types, which collectively account for over 67\% of the dataset. 

\begin{table}[htbp]
  \centering
  \begin{tabular}{@{}lrr@{}}
    \toprule
    Error Type & Count & Percentage \\
    \midrule
    M:DET &  26,008 & 19.0\% \\
    R:NOUN:NUM &  21,149 & 15.5\% \\
    R:PREP & 18,702 & 13.7\% \\
    U:DET & 15,708 & 11.5\% \\
    R:VERB:TENSE & 10,599 & 7.7\% \\
    \bottomrule
  \end{tabular}
  \caption{Distribution of the top 5 ERRANT error types in BLiSS.}
  \label{tab:errant-top5-bliss-precise}
\end{table}

\paragraph{Learner Demographics}
The rich metadata from the source corpora allows for detailed analysis across learner populations. 
Table~\ref{tab:l1-top5-bliss-precise} shows the distribution of the top five L1 backgrounds in the dataset. 
The significant representation of typologically diverse languages such as Chinese, Japanese, and Arabic makes the benchmark particularly powerful for investigating L1 transfer effects. 
\begin{table}[htbp]
  \centering
  \begin{tabular}{@{}lrr@{}}
    \toprule
    L1 Background & Count & Percentage \\
    \midrule
    Chinese & 23,771 & 17.4\% \\
    Japanese & 14,478 & 10.6\% \\
    Italian & 11,918 & 8.7\% \\
    French & 11,486 & 8.4\% \\
    Arabic & 9,484 & 6.9\% \\
    \bottomrule
  \end{tabular}
  \caption{Distribution of the top 5 L1 backgrounds in BLiSS.}
  \label{tab:l1-top5-bliss-precise}
\end{table}

\begin{figure}[htbp]
\centering
\begin{tikzpicture}
\begin{axis}[
  height=4.8cm,
  width=\columnwidth,
  ybar stacked,
  bar width=12pt,
  xlabel={CEFR Level},
  ylabel={Triplets},
  ymin=0, ymax=50000,
  scaled y ticks=false,
  ytick distance=10000,
  y tick label style={/pgf/number format/1000 sep=,},
  tick align=inside,
  ymajorgrids=true,
  grid style=dashed,
  axis x line*=bottom,
  axis y line*=left,
  tickwidth=0pt,
  enlarge x limits=0.12,
  label style={font=\footnotesize},
  tick label style={font=\scriptsize},
  xticklabel style={rotate=35, anchor=east},
  symbolic x coords={A1, A2, B1, B2, C1, C2},
  xtick=data,
  legend style={
    at={(0.5,1.22)},
    anchor=north,
    legend columns=3,
    /tikz/every even column/.append style={column sep=0.3cm}
  }
]
\addplot+[fill=red!60] coordinates {
  (A1, 39)
  (A2, 1994)
  (B1, 7771)
  (B2, 6154)
  (C1, 1397)
  (C2, 25)
};
\addlegendentry{Write \& Improve}

\addplot+[fill=green!60] coordinates {
  (A1, 0)
  (A2, 0)
  (B1, 3394)
  (B2, 10214)
  (C1, 0)
  (C2, 0)
};
\addlegendentry{FCE}

\addplot+[fill=blue!60] coordinates {
  (A1, 29044)
  (A2, 15583)
  (B1, 22223)
  (B2, 30379)
  (C1, 7804)
  (C2, 0)
};
\addlegendentry{EFCamDat}
\end{axis}
\end{tikzpicture}
\caption{Distribution of CEFR proficiency levels in BLiSS by corpus (stacked triplet counts).}
\label{fig:cefr-stacked-by-corpus}
\end{figure}
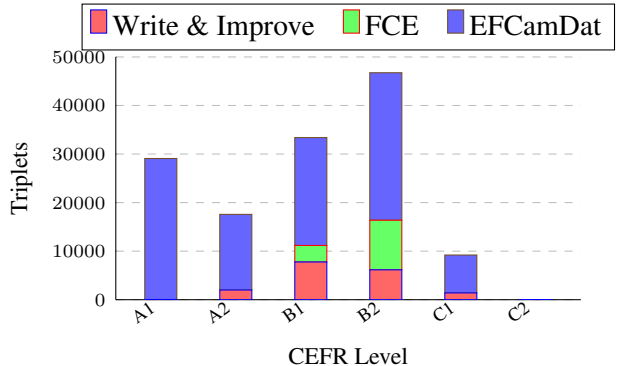

In terms of learner proficiency, BLiSS spans a wide range of the CEFR scale, from beginner (A1) to advanced (C2), as detailed in Figure~\ref{fig:cefr-stacked-by-corpus}. 
The dataset has substantial representation between the beginner and intermediate(A1 - B2) levels but significantly less at higher levels with only 25 triplets at the C2 level. 
This broad distribution is a key strength, enabling the study of how model behavior might differ when evaluated on errors typical of different proficiency levels.

\section{Evaluation}

The core objective is to quantitatively measure a model's alignment to the naturalistic production errors produced by second language (L2) learners of English. To evaluate a model’s selective tolerance, we introduce a set of complementary metrics that capture different aspects of its behavior. The \textbf{Learner Preference (LP)} metric provides a simple metric that measures whether the model prefers a human learner sentence over the corrected version, though a high LP could reflect either accurate simulation of learner tendencies or poor grammatical knowledge. To directly probe selective tolerance, \textbf{Human vs. Artificial Preference (HAP)} measures whether the model favors naturalistic learner errors over contrived, artificial errors, while \textbf{HAP-\(\tau\)} is a stricter version that ensures the model’s preference is meaningful and not just due to numerical noise. Finally, the \textbf{Strict Order (SO)} metric captures the most stringent behavior, requiring the model to rank all three sentences in the hypothesized order—corrected first, learner second, artificial last—indicating a balance between grammatical competence and nuanced sensitivity to L2 error patterns. Together, these metrics provide a multi-faceted view of whether a language model can recognize correct grammar, differentiates between plausible and implausible errors, and exhibits robust, cognitively plausible error sensitivity.

A model’s preference for a sentence is quantified using token-normalized surprisal, measured in Bits Per Token (BPT), where low BPT indicates high plausibility under the model’s learned distribution and high BPT signals a grammatical deviation. By computing BPT scores for each sentence in a BLiSS triplet—including the corrected sentence, the human learner error, and an artificially generated error—we can evaluate not only whether a model recognizes correct grammar, but also whether it differentiates between naturalistic learner errors and contrived mistakes. These BPT scores underpin the three evaluation metrics in the BLiSS framework. 

We recommend that  \textbf{each metric should be reported separately}, as they provide complementary insights: \textsc{Learner Preference} (LP) captures general grammatical preference, \textsc{Human  v Artificial Preference} (HAP and HAP-\(\tau\)) metrics assess selective tolerance, and \textsc{Strict Order} (SO) evaluates the full hypothesized ranking. Combining these metrics into a single score would obscure these distinctions and reduce the interpretability of a language model’s behavior on L2 error patterns.

% To evaluate if a model is truely 'learner-like', we need a framework that goes beyond standard grammaticality. 
% Out evaluation is designed as a series of 'tests' to probe a model's internal grammar, asking not just 'Is this sentence correct?' but 'does this model show sensitivity to the patterns of a human learner?'
% To assess a model's selective tolerance for learner errors, we employ an evaluation framework based on token-normalized surprisal. 
% This framework is designed to answer two primary questions:
% \begin{enumerate}
%     \item To what extent does a model penalize an naturalistic human error less than a matched, artificial-locus error?
%     \item Is this behaviour statistically significant and is it sensitive to the learner's L1 background?
% \end{enumerate}

\subsection{Scoring Signal}
We quantify a model's preference for a given sentence $s$ by its token-normalized surprisal, measured in Bits Per Token (BPT). This is calculated as the negative log-likelihood of the sentence, normalized by the number of tokens. 
$$BPT(s) = -\frac{1}{|s|}\sum_{t=1}^{|s|}log_2p(w_t|w_{<t})$$
where $|s|$ is the number of tokens in the sentence and $p(w_t|w_{<t})$ is the probability assigned by the model to token $w_t$ given the proceeding context. 

From a cognitive perspective, surprisal is often used as a proxy for processing effort. 
A sentence that aligns with a model's learned grammatical and statistical patterns will have low surprisal (low BPT), indicating it is highly plausible under the model's distribution. 
Conversely, a sentence with a grammatical deviation will have high surprisal (high BPT). 
This allows us to use BPT as a 'plausibility score' to measure the model's preference for each of the three sentences in a BLiSS triplet. 

For each item in the BLiSS dataset, we apply this scoring signal to three sentences in the triplet, which we will formally denote as: $s_{corr}$ (corrected), $s_{lrn}$ (learner), and $s_{art}$ (artificial). 
% \subsection{Quantifying Selective Tolerance}
% For each triplet in the BLiSS dataset $(corr, lrn, art)$, we use the BPT scores to calculate the penalty the model assigns to each error relative to the corrected sentence:
% $$\Delta_{learner} = BPT(lrn) - BPT(corr)$$
% $$\Delta_{random} = BPT(art) - BPT(corr$$
% The core measure of selective tolerance is the margin $m$, defined as the difference between these two penalties:
% $$m = \Delta_{random} - \Delta_{learner}$$
% A positive margin ($m > 0$) indicates that the penalty for the artificial error is greater than the penalty for the human error. 
% This is the primary signal of the desired behaviour: evidence of selective tolerance. 

\subsection{Evaluation Metrics}
We present a suite of metrics designed to provide a multi-faceted view of a model's behavior. 
These metrics are organized around two key concepts: a baseline measure of simple learner preference and our primary measures of selective tolerance. 

\paragraph{Baseline Metric: Learner Preference (LP)}
We provide this metric as a minimal-pair evaluation. 
We define Learner Preference (LP) as the proportion of items where the model prefers the learner sentence over the corrected version: $BPT(s_{lrn}) < BPT(s_{corr})$. 
The motivation for LP is that for certain applications, such as simulating learner output, a model might be intentionally designed to reproduce learner errors. 
However, LP is inherently ambiguous, as a high score could also simply reflect poor grammatical knowledge. 
We therefore use it as a diagnostic baseline. 

\paragraph{Selective Tolerance Metrics}
To overcome the ambiguity of LP, our primary metrics are designed to probe a model's selective tolerance directly. 
The desired behavior for a cognitively plausible model is twofold: it should, first and foremost, still recognize and prefer correct grammar, yet it should also differentiate between the plausibility of different types of errors. 
Specifically, it should find a naturalistic, systematic learner error to be more plausible (less surprising) than a contrived, artificial error. 
According to this principle, the ideal ordering of preferences for any triplet should be the corrected sentence, followed by the learner sentence, and finally the artificial sentence. 
Following this, we present three primary metrics that quantify a model's adherence to this behavior. 

\begin{enumerate}
    \item \textbf{SO (Strict Order):} This is the most stringent metric. It measures the proportion of the items where the model's preferences follow the full, hypothesized order of plausibility: $BPT(s_{corr}) < BPT(s_{lrn}) < BPT(s_{art})$. A high SO score is the strongest evidence that a model successfully balances grammatical competence with a nuanced sensitivity to interlanguage. 
    \item \textbf{HAP (Human vs. Artificial Preference):} This metric isolates the central test of selective tolerance by measuring the proportion of items where the model simply prefers the human error over the artificial one: $BPT(s_{\text{lrn}}) < BPT(s_{\text{art}})$. HAP allows us to credit a model for correctly distinguishing between the two error types. 
    \item \textbf{HAP-$\tau$ (Robust HAP):} A stricter version of HAP, this metric requires the BPT difference between the artificial and learner sentences to exceed a small positive buffer $\tau$: $BPT(s_{\text{art}}) - BPT(s_{\text{lrn}}) > \tau$. This ensures the model's preference is confident and meaningful, rather than an artifact of numerical noise. 
\end{enumerate}

\section{Models}
\label{sec:models}
We evaluate a diverse range of models on the BLiSS benchmark. 
The models are grouped into four distinct families, ordered by their increasing degree of specialization for second language acquisition (SLA). 
This progression allows us to systematically investigate how training data, architecture, and SLA-inspired objectives influence a model's capacity for selective tolerance. 

\paragraph{Standard Bilingual LLMs}
This family serves as our baseline, representing powerful, general-purpose models that have not been specifically designed to model learner language or the acquisition process. 
These are large language models trained on massive corpora of standard, native-speaker text in two languages. Their training objective is to model fluent, grammatical language, not the intermediate stages of learning. 
We include Bilingual-GPT-NeoX-4B\footnote{\url{https://huggingface.co/rinna/bilingual-gpt-neox-4b}}   (Japanese--English) \citep{rinna-bilingual-gpt-neox-4b, sawada2024release}, CroissantLLM\footnote{\url{https://huggingface.co/croissantllm/CroissantLLMBase}} \citep{Faysse2024CroissantLLMAT} (French--English), and MAP-Neo-7B\footnote{\url{https://map-neo.github.io/}} \citep{zhang2024mapneo} (Chinese--English).

\paragraph{Bilingual BabyLMs}
This family represents models that are 'acquisition-inspired' in their data scale but are not explicitly designed for SLA. 
These are smaller models trained from scratch on developmentally plausible, child-directed speech (CDS) in two languages. While they model the acquisition of language, they are primarily simultaneous bilingual first language acquisition (BFLA), not successive L2 learning. 
We evaluate publicly released models \citep{jumelet2025babybabellm}\footnote{\url{https://huggingface.co/BabyLM-community}} trained from scratch on 10M words of CDS in English plus one other language (Persian, German, Indonesian, Japanese, Dutch, or Chinese).

\paragraph{Acquisition-Inspired L2 Models}
This family includes models that explicitly incorporate principles from SLA research into their design. 
They are designed to simulate the process of an L1 speaker learning an L2, often through sequential training regimes or other architectural priors that model transfer. 
SLABERT \citep{yadavalli-etal-2023-slabert} follows the \emph{Test for Inductive Bias via Language Model Transfer} \citep[TILT;][]{pauls-klein-2012-large}: pretrain on age-ordered CDS in L1 (French, Polish, Indonesian, Japanese), then fine-tune on English adult-directed speech with all parameters frozen except embeddings.  
B-GPT \citep{Arnett2025OnTA} is trained with \emph{sequential} exposure (L1 then L2) or \emph{simultaneous} exposure (L1+L2 mixed), here evaluated only for English L2 with Dutch or Spanish L1.

\paragraph{Learner-Trained Models}
This final family represents models that are directly exposed to learner language during training. 
Instead of learning from native text and hoping learner-like patterns emerge, we train these models on the same kind of data used in our benchmark. 
We train several GPT-2 medium models from scratch on learner-produced English essays from the Cambridge Learner Corpus (CLC) and EFCAMDAT.
To ensure fairness in evaluation, these models are only evaluated on the W\&I slice of BLiSS.

\section{Results}
\begin{table*}[h!]
\centering
\renewcommand{\arraystretch}{1.15} % improves readability
\setlength{\tabcolsep}{3pt} % reduces horizontal padding
\begin{tabular}{l *{8}{p{1.2cm}}}
\toprule
\multirow{3}{*}{Model} &
\multicolumn{7}{c}{BLiSS} &
\multirow{3}{*}{BLiMP} \\
\cmidrule(lr){2-8}
& \multicolumn{2}{c}{HAP} 
& \multicolumn{2}{c}{HAP@$\tau$} 
& \multicolumn{2}{c}{SO} 
& \multicolumn{1}{c}{LP} 
& \\
\cmidrule(lr){2-3}\cmidrule(lr){4-5}\cmidrule(lr){6-7}
& Overall & L1 & Overall & L1 & Overall & L1 &  & \\
\midrule
\multicolumn{9}{l}{\textit{Bilingual LLMs}} \\
CroissantLLM & 67.51* & 81.76** & 57.42* & 71.70**  & 57.64*  & 54.09 & 12.84 & 81.20 \\
Neox-4B      & 60.55* & 78.26** & 42.98* & 62.32**  & 35.15*  & 1.88 & 16.27 & 82.12 \\
MAP-Neo-7B   & 66.81* & 77.12 & 58.14* & 72.03** & 56.05* & 45.76** & 14.14 & 82.12 \\
\midrule\midrule
\multicolumn{9}{l}{\textit{Bilingual BabyLM models}} \\
BBLM--DE & 60.15* & 76.92** & 50.59* & 66.67** & 43.73* & 33.33 & 18.80 & 66.32 \\
BBLM--ZH & 59.56* & 72.88** & 49.57* & 66.95** & 34.93* & 38.14 & 20.44 & 66.44 \\
BBLM--ID & 60.08* & 66.60 & 50.41* & 62.96 & 43.66* & 37.04 & 28.57 & 66.22 \\
BBLM--FR & 60.38* & 79.25** & 50.70* & 67.92** & 43.93* & 44.03 & 13.71 & 66.10 \\
\midrule\midrule
\multicolumn{9}{l}{\textit{SLABERT}} \\
SLABERT--JP & 50.42* & 47.83 & 31.40* & 27.54 & 16.58* & 15.94 & 63.46 & 49.16 \\
SLABERT--FR & 52.22* & 52.20 & 34.50* & 33.96 & 16.20* & 15.09 & 57.47 & 48.44 \\
SLABERT--ID & 46.36* & 38.89 & 30.91* & 25.93 & 15.40* & 12.96 & 57.14 & 51.36 \\
SLABERT--PL & 46.01* & 54.92** & 32.99* & 38.52 & 14.07* & 18.85 & 57.57 & 52.00 \\
\midrule
\multicolumn{9}{l}{\textit{B-GPT}} \\
B-GPT-ES-SIM & 66.43* & 77.14** & 56.48* & 62.14** & 54.57* & 50.00 & 12.99 & 52.66 \\
B-GPT-ES-SEQ & 66.06* & 74.29** & 56.15* & 61.43** & 55.06 & 48.57** & 12.17 & 54.19 \\
\midrule\midrule
\multicolumn{9}{l}{\textit{EFCAMDAT Trained}} \\
LM--EF    & 51.87* & 47.94** & 36.94* & 33.78** & 15.23* & 12.66** & 39.51 & 53.76 \\
Noise--EF & 47.69* & 46.26 & 28.47* & 29.67 & 11.40* & 11.33 & 41.95 & 54.84 \\
Contr--EF & 62.09* & 62.24 & 48.02* & 41.74 & 23.70* & 18.03 & 69.51 & 50.04 \\
Compl--EF & 49.50* & 56.23 & 44.75* & 45.80 & 21.35* & 19.54 & 40.73 & 54.74 \\
\bottomrule
\end{tabular}
\caption{BLiSS metrics (HAP, HAP@$\tau$, Strict Order, LP) with L1 and Overall subcolumns, alongside BLiMP grammaticality accuracy. 
An asterisk (*) indicates performance significantly above the 50\% chance baseline ($p < 0.05$), while a double asterisk (**) on L1 scores indicates a statistically significant difference between the L1-specific and overall performance. 
See full result table in Section~\ref{sec:full_result}.}
\label{tab:bliss_cefr_ll}
\end{table*}

Table~\ref{tab:bliss_cefr_ll} presents BLiSS scores for all evaluated models. 
Our analysis shows three primary findings that validate the BLiSS benchmark as a tool for measuring a distinct, acquisition-related dimension of model behavior. 

An analysis of the model families in Table \ref{tab:bliss_cefr_ll} reveals distinct performance profiles. The Bilingual LLMs and B-GPT models emerge as the strongest performers on our primary selective tolerance metrics. Both families form tight clusters with high HAP scores ($\approx$66-67\%) and, notably, the highest Strict Order (SO) scores ($\approx$55-57\%). This indicates a robust ability to correctly rank the full triplet.

The Bilingual BabyLM models also perform significantly above chance, but with lower SO scores ($\approx$35-44\%), suggesting a weaker, though still present, signal of selective tolerance. A consistent and important trend among these three families is a statistically significant increase in performance on their respective L1 data slices, providing strong evidence that they have internalized L1-dependent transfer patterns and validating BLiSS as a tool for probing these fine-grained behaviors.

In contrast, the SLABERT and Learner-Trained models show a different and less successful profile. Their very high Learner Preference (LP) scores (often >50\%) are coupled with poor performance on our primary selective tolerance metrics, particularly Strict Order. This suggests that their training may have made them indiscriminately accepting of learner-like forms, hindering their ability to distinguish between plausible human errors and implausible artificial ones.

\subsection{BLiSS vs. BLiMP}
\begin{figure}[htbp]
\centering
\begin{tikzpicture}
\begin{axis}[
    height=7cm,
    width=0.9\columnwidth,
    xlabel={Selective Tolerance: HAP},
    ylabel={Grammaticality: BLiMP Score},
    xmin=40, xmax=80,
    ymin=45, ymax=90,
    title={Selective Tolerance vs. Grammaticality},
    legend style={
        at={(0.5, -0.25)},
        anchor=north,
        legend columns=2,  
        /tikz/every even column/.append style={column sep=0.5cm}
    },
    grid=major,
    grid style=dashed,
    scatter/classes={
        a={mark=*, red},         % Bilingual LLMs
        b={mark=square*, blue},  % Bilingual BabyLMs
        c={mark=triangle*, orange}, % SLABERT
        d={mark=diamond*, green!60!black}, % B-GPT
        e={mark=*, purple},     % EFCAMDAT Trained
        f={mark=square*, brown}   % CLC Trained
    }
]

\addplot[scatter, only marks, scatter src=explicit symbolic]
table [meta=label] {
    x       y       label
    % Bilingual LLMs
    67.51   81.20   a
    60.55  82.12   a
    66.81   82.12   a
    
    % Bilingual BabyLMs
    60.15   66.32   b
    59.56   66.44   b
    60.08   66.22   b
    60.38   66.10

    % SLABERT
    50.42   49.16   c
    52.22   48.44   c
    46.36   51.36   c
    46.01   52.00   c

    % B-GPT
    66.43   52.66   d
    66.06   54.19   d
    66.17   57.50   d
    66.28   54.32   d

    % EFCAMDAT Trained
    51.87   53.76   e
    47.69   54.84   e
    62.09   50.04   e
    49.50   54.74   e

    % CLC Trained
    61.94   54.68   f
    58.00   54.55   f
    61.27   53.79   f
    55.18   54.95   f
    51.89   52.70   f
};

\legend{
    Bilingual LLMs,
    Bilingual BabyLMs,
    SLABERT,
    B-GPT,
    EFCAMDAT Trained,
    CLC Trained
}

\end{axis}
\end{tikzpicture}
\caption{Selective tolerance (HAP score) versus grammaticality (BLiMP score) across all evaluated models. Each point represents a model, colour-coded by its training family. }
\label{fig:hap_vs_blimp}
\end{figure}
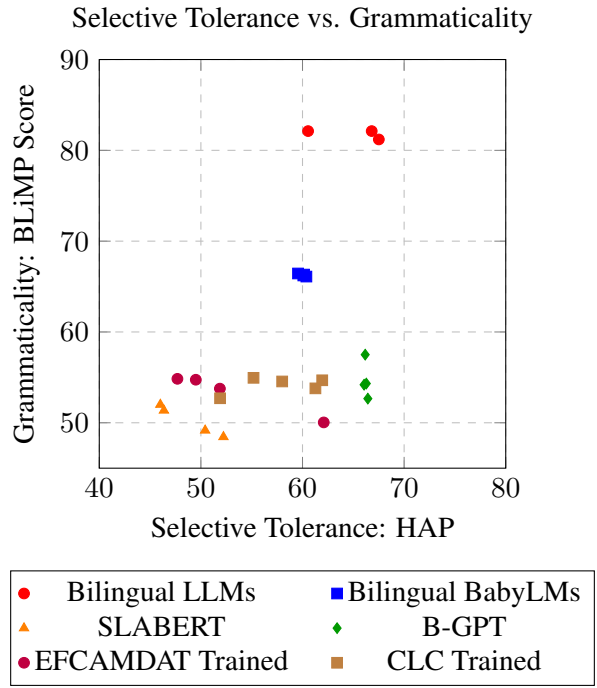

To visualize the relationship between a model's BLiSS and BLiMP scores, Figure~\ref{fig:hap_vs_blimp} plots the HAP score against the BLiMP score for each evaluated mode, colour-coded by the model family. 
The plot demonstrates several key insights into the nature of the BLiSS benchmark and the capabilities of different model architectures. 

A striking observation is that models from the same training family form tight clusters. 
For example, the large Bilingual LLMs occupy a distinct region in the top-right of the plot, while the B-GPT and SLABERT models form their own clear groups. 
This consistency is a powerful validation of our methodology; it suggests that BLiSS is successfully capturing a stable signal that is reflective of the underlying training paradigm, rather than just idiosyncratic model behavior. 
The two learner-trained families (EFCAMDAT and CLC) show slightly more internal variance, which is expected, as the primary differentiating factor within those families is the training data. 

Another pattern we observe from the plot is the clear lack of a strong positive correlation between the two metrics. 
High performance on BLiMP does not guarantee high performance on BLiSS and vice-versa. 
The large Bilingual LLMs, for instance, excel at both. 
However, other models achieve strong selective tolerance without top-tier grammaticality. 
The B-GPT models are a prime example. 

This demonstrates that BLiSS offers a complementary, second dimension for language model evaluation. It measures a distinct capability that is not captured by standard grammaticality benchmark alone.

\section{Conclusion}
\label{sec:conclusion}

As the BabyLM Challenge extends cognitively-inspired language modeling beyond English, there are methodological challenges in evaluating the formal competence of BabyLM-inspired L2LMs that are modeling second language or bilingual acquisition. 
To address this, we introduced BLiSS, a large-scale benchmark built on a new paradigm of selective tolerance. By evaluating models on controlled triplets (corrected, learner error, artificial error), BLiSS measures a model's ability to distinguish naturalistic human errors from contrived ones, disentangling sensitivity to learner patterns from general grammatical competence. Our experiments demonstrate that selective tolerance is a distinct capability from standard grammaticality, with performance clustering strongly by training paradigm and revealing sensitivity to L1-specific transfer effects. We hope that BLiSS will serve as both a benchmark and a research catalyst for developing L2 language models that better reflect the diversity and systematicity of human language acquisition.
% We have introduced BLiSS, a large-scale naturalistic benchmark derived from parallel learner corpora for evaluating the learner-likeness of language models across proficiency levels and L1 backgrounds. 

% BLiSS complements grammaticality-oriented evaluations such as BLiMP by operationalising learner-likeness as a model's preference for naturalistic learner productions over their corrected forms. The dataset aggregates over 1.5M sentence--correction pairs from three major learner corpora, enriched with ERRANT annotations and organised into CEFR- and L1-specific subsets.

% Our experiments show that standard bilingual models trained only on native text fail to capture learner-like preferences, whereas SLA-inspired training regimes—particularly those incorporating sequential L1$\rightarrow$L2 exposure or contrastive objectives—yield substantially higher BLiSS scores without uniformly degrading grammaticality. BLiSS also supports fine-grained probing of specific error types with well-established L1 dependencies, recovering expected patterns such as persistent article difficulty for Japanese learners.

% We hope that BLiSS will serve as both a benchmark and a research catalyst for developing cognitively-informed L2 language models that reflect the diversity and systematicity of learner language.

% Add these to your document's preamble if they are not already there:
% \usepackage{tikz}
% \usepackage{pgfplots}
% \pgfplotsset{compat=1.17} % Use a recent version

% \todo{I think a short summary is needed to end the paper: remind the reader of the contributions}

\section*{Limitations}

Several limitations should be considered when interpreting our results. First, BLiSS relies on sentence-level corrections from learner corpora, which may not capture all aspects of learner language development. The benchmark focuses on grammatical and lexical errors but does not assess discourse-level phenomena, pragmatic competence, or other dimensions of L2 proficiency that extend beyond sentence boundaries.
This imbalance may affect the reliability of conclusions about advanced learner behavior and limits our ability to study developmental trajectories at higher proficiency levels.

Specific L1 backgrounds and grammatical error types that were already infrequent in the source corpora become even more sparse in the final dataset. The low success rate of our generation process (4.8\%) means that only the most common and structurally regular phenomena are represented at scale. This may limit the statistical power for fine-grained analyses on these lower-frequency L1-error combinations and means that our results are most representative of common error patterns.

\section*{Acknowledgments}

With thanks to Laura Barbenel for proof-reading this manuscript. We thank the anonymous reviewers for their useful feedback and suggestions, which greatly improved the manuscript.  This paper reports on work supported by Cambridge University Press \& Assessment. 

% for HPC?
%It was performed using resources provided by the Cambridge Service for Data Driven Discovery (CSD3) operated by the University of Cambridge Research Computing Service, provided by Dell EMC and Intel using Tier-2 funding from the Engineering and Physical Sciences Research Council (capital grant EP/T022159/1), and DiRAC funding from the Science and Technology Facilities Council.
% only required if we used bubbles?
%Additionally, we thank the NVIDIA Corporation for the donation of the Titan X Pascal GPU used in this research.

%\bibliographystyle{acl_natbib}
\bibliography{custom}
\appendix
\onecolumn

\section{Full Evaluation Results}
\label{sec:full_result}
\begin{table*}[h!]
\centering
\begin{tabular}{l *{8}{p{1.2cm}}}
\toprule
\multirow{3}{*}{Model} &
\multicolumn{7}{c}{BLiSS} &
\multirow{3}{*}{BLiMP} \\
\cmidrule(lr){2-8}
& \multicolumn{2}{c}{HAP} & \multicolumn{2}{c}{HAP@$\tau$} & \multicolumn{2}{c}{SO} & \multicolumn{1}{c}{LP} & \\
\cmidrule(lr){2-3}\cmidrule(lr){4-5}\cmidrule(lr){6-7}
& Overall & L1 & Overall & L1 & Overall & L1 &  & \\
\midrule
\multicolumn{9}{l}{\textit{Bilingual LLMs}} \\
CroissantLLM & 67.51* & 81.76** & 57.42* & 71.70**  & 57.64*  & 54.09 & 12.84 & 81.20 \\
Neox-4B      & 60.55* & 78.26**   & 42.98* & 62.32**  & 35.15*  & 1.88 & 16.27& 82.12 \\
MAP-Neo-7B   & 66.81* & 77.12   & 58.14* & 72.03**     & 56.05*  & 45.76** & 14.14& 82.12 \\
\midrule\midrule
\multicolumn{9}{l}{\textit{Bilingual BabyLM models}} \\
BBLM--DE & 60.15* & 76.92**   & 50.59* & 66.67** & 43.73* & 33.33   & 18.80 & 66.32 \\
BBLM--ZH &59.56* & 72.88**   & 49.57* & 66.95**   & 34.93* & 38.14 & 20.44 & 66.44 \\
% BBLM--FA &  &  &  &  &  &  & — & 65.84 \\
BBLM--ID & 60.08* & 66.6 & 50.41* &62.96 & 43.66* & 37.04 & 28.57 & 66.22 \\
% BBLM--NL & 65.21* & 67.12   & 53.01* & 56.08   & 34.98* & 40.08** & 14.75 & 68.24 \\
BBLM--FR & 60.38* & 79.25** & 50.70* & 67.92** & 43.93* & 44.03   & 13.71 &  66.10\\
\midrule\midrule
\multicolumn{9}{l}{\textit{SLABERT}} \\
% SLABERT--DE &  &  &  &  &  &  & — & 52.33 \\
SLABERT--JP & 50.42* & 47.83 & 31.40* & 27.54 &16.58* & 15.94   & 63.46 & 49.16 \\
SLABERT--FR & 52.22* & 52.20   & 34.50* & 33.96 & 16.20* & 15.09 & 57.47 & 48.44 \\
SLABERT--ID & 46.36* & 38.89 & 30.91* & 25.93 & 15.40* & 12.96   & 57.14 & 51.36 \\
SLABERT--PL & 46.01* & 54.92** & 32.99* & 38.52 & 14.07* & 18.85   & 57.57 & 52.00 \\
% SLABERT--DE & 53.40* & 53.29   & 30.20* & 30.78 & 12.76* & 11.57   & — &  \\
\midrule
\multicolumn{9}{l}{\textit{B-GPT}} \\
B-GPT-ES-SIM & 66.43* & 77.14**  & 56.48 * & 62.14** & 54.57* & 50.00   & 12.99 & 52.66 \\
B-GPT-ES-SEQ & 66.06* & 74.29** & 56.15* & 61.43** & 55.06  & 48.57**   & 12.17 & 54.19 \\
% B-GPT-NL-SIM & 66.17* & - & 56.31* & - & 54.70* & - & 50.82 & 57.50 \\
% B-GPT-NL-SEQ & 66.28* & - & 56.46* & - & 55.60* & - & 44.26 & 54.32 \\
\midrule\midrule
\multicolumn{9}{l}{\textit{EFCAMDAT Trained}} \\
LM--EF    & 51.87* & 47.94** & 36.94* & 33.78** & 15.23* & 12.66** & 39.51 & 53.76 \\
Noise--EF & 47.69*  & 46.26   & 28.47*  & 29.67   & 11.40*  & 11.33   & 41.95 & 54.84 \\
Contr--EF & 62.09*  & 62.24   & 48.02*  & 41.74   & 23.70*  & 18.03   & 69.51 & 50.04 \\
Compl--EF & 49.50*  & 56.23   & 44.75*  & 45.80   & 21.35*  & 19.54   & 40.73 & 54.74 \\
\midrule
\multicolumn{9}{l}{\textit{CLC Trained}} \\
CLC--A1 & 61.94* &  - & 51.94* &  - & 28.86* &  - & 35.28 & 54.68 \\
CLC--A2 & 58.00* &  - & 46.94* & -  & 23.74* &  - & 39.27 & 54.55 \\
CLC--B1 & 61.27*&  - & 50.75* &  - & 29.62*& -  & 29.62 & 53.79 \\
CLC--B2 & 55.18* &  - & 43.47* & -  & 21.93* &  - & 41.68 & 54.95 \\
CLC--C1 & 51.89* &  - & 38.65* &  - & 18.48* &  - & 46.02 & 52.70 \\
\bottomrule
\end{tabular}
\caption{BLiSS metrics (HAP, HAP@$\tau$, Strict Order, LP) with L1 and Overall subcolumns, alongside BLiMP grammaticality accuracy. An asterisk (*) indicates performance significantly above the 50\% chance baseline (p < 0.05), while a double asterisk (**) on L1 scores indicates a statistically significant difference between the L1-specific and overall performance.}
\label{tab:bliss_full}
\end{table*}
\newpage 

\section{Learner-Trained Model Details}
All models were trained using the HuggingFace \texttt{Trainer} API with the following configuration.  
Training ran for 10 epochs for CLC-trained models and 5 epochs for EFCAMDAT-trained models. 
\begin{table}[h]
\centering
\begin{tabular}{ll}
\toprule
\textbf{Parameter} & \textbf{Value} \\
\midrule
Seed & 42 \\
Block size & 1024 tokens \\
Per-device batch size & 2 \\
Gradient acc. steps & 8 \\
Effective batch size & 16 \\
Learning rate & $5\times 10^{-5}$ \\
Weight decay & 0.1 \\
Warmup steps & 500 \\
Logging steps & 50 \\
Max steps & $-1$ (full epochs) \\
Scheduler & cosine \\
Optimiser & AdamW\\
Mixed precision & \texttt{fp16}\\
Gradient checkpointing & Enabled \\
Save strategy & End of each epoch \\
\bottomrule
\end{tabular}
\caption{Training hyperparameters.}
\label{tab:training_hparams}
\end{table}

\newpage 

\section{ERRANT Annotation Scheme} \label{errant}

All learner sentences in BLiSS are automatically annotated with ERRANT v3.0.0 to obtain token-level error labels.

\begin{table}[ht]
\centering
\caption{Complete list of valid error code combinations}
\begin{tabular}{|l|l|l|l|l|}
\hline
\textbf{Operation Tier} & \textbf{Type} & \textbf{Missing} & \textbf{Unnecessary} & \textbf{Replacement} \\ \hline
\multirow{11}{*}{Token Tier} 
& Adjective      & M:ADJ & U:ADJ & R:ADJ \\ 
& Adverb         & M:ADV & U:ADV & R:ADV \\
& Conjunction    & M:CONJ & U:CONJ & R:CONJ \\
& Determiner     & M:DET & U:DET & R:DET \\
& Noun           & M:NOUN & U:NOUN & R:NOUN \\
& Particle       & M:PART & U:PART & R:PART \\
& Preposition    & M:PREP & U:PREP & R:PREP \\
& Pronoun        & M:PRON & U:PRON & R:PRON \\
& Punctuation    & M:PUNCT & U:PUNCT & R:PUNCT \\
& Verb           & M:VERB & U:VERB & R:VERB \\ \cline{2-5}
& Other          & M:CONTR & U:CONTR & R:CONTR \\
& Morphology     & -      & -      & R:MORPH \\
& Orthography    & -      & -      & R:ORTH \\
& Other          & M:OTHER & U:OTHER & R:OTHER \\
& Spelling       & -      & -      & R:SPELL \\
& Word Order     & -      & -      & R:WO \\ \hline
\multirow{9}{*}{Morphology Tier}
& Adjective Form     & -      & -      & R:ADJ:FORM \\
& Noun Inflection    & -      & -      & R:NOUN:INFL \\
& Noun Number        & -      & -      & R:NOUN:NUM \\
& Noun Possessive    & M:NOUN:POSS & U:NOUN:POSS & R:NOUN:POSS \\
& Verb Form          & M:VERB:FORM & U:VERB:FORM & R:VERB:FORM \\
& Verb Inflection    & -      & -      & R:VERB:INFL \\
& Verb Agreement     & -      & -      & R:VERB:SVA \\
& Verb Tense         & M:VERB:TENSE & U:VERB:TENSE & R:VERB:TENSE \\ \hline
\end{tabular}
\end{table}

\newpage 

%\section{ERRANT Error Analysis of BLiSS}\label{errant-error-analysis}

% We provide in Table~\ref{tab:errant_top10} a list of the ten most frequent error types across the combined dataset. 

%\begin{table}[h]
%\centering
%\caption{Top 10 ERRANT error types across BLiSS.}
%\label{tab:errant_top10}
%\begin{tabular}{lrr}
%\toprule
%Error type & Count & Proportion \\
%\midrule
%R:SPELL        & 508,929 & 17.60\% \\
%R:OTHER        & 315,267 & 10.90\% \\
%M:DET          & 172,417 & 5.96\% \\
%R:PREP         & 129,106 & 4.46\% \\
%R:NOUN:NUM     & 117,290 & 4.06\% \\
%R:NOUN         & 111,963 & 3.87\% \\
%R:VERB         & 106,136 & 3.67\% \\
%U:DET          & 102,273 & 3.54\% \\
%R:VERB:TENSE   & 88,322  & 3.05\% \\
%R:VERB:SVA     & 82,762  & 2.86\% \\
%\bottomrule
%\end{tabular}
%\end{table}

\end{document}